\documentclass[sigconf,nonacm]{acmart}

\usepackage{booktabs}
\usepackage{graphicx}

\AtBeginDocument{%
  }

\makeatletter
\newcommand{\acmnocountry}{\@ifundefined{@ACM@countrypresenttrue}{}%
  {\global\@ACM@countrypresenttrue}}
\makeatother

\begin{document}

\title{How Execution Assumptions Change Short-Horizon Sharpe Rankings:
Evidence from a Synthetic Trading Benchmark}

\author{Weicheng Xue}
\affiliation{\institution{Virginia Tech}\acmnocountry}
\email{weich97@vt.edu}

\begin{abstract}
Backtests of LLM trading agents often assume that every order fills at the
closing price. We ask whether this choice changes only reported returns or also
the order of the agents. Five prompted LLM signal policies and seven classical
baselines trade the same synthetic price paths under six execution settings,
from near-ideal fills to latency, spread, participation, and impact stresses.
The main experiment contains $2{,}462$ runs with matched decision frequencies
and paired market paths. On the compressed two-asset board, agreement between
the near-ideal and default-stress rankings falls to Kendall $\tau_b=0.21$ in the
high-volatility regime, compared with $0.82$ in the calm regime. The
seed-bootstrap intervals, $[0.00,0.52]$ and $[0.48,0.94]$, are wide and overlap.
On a fixed 11-policy board, agreement rises from 0.24 with two assets to 0.85
with ten; the two-asset point estimate differs substantially from the wider
settings we tested. Rank changes are related to turnover, and comparisons with
buy-and-hold also depend on how that anchor is initialized. The experiment does
not compare LLM trading skill. It shows that, on a short horizon, an execution
convention can become part of the benchmark's headline. Execution assumptions
and rank stability should be reported alongside returns.
\end{abstract}



\maketitle
\hypersetup{pdfauthor={Weicheng Xue}}

\section{Introduction}

LLM trading agents are often ranked by backtested returns. The model observes a
market state, proposes a trade, and receives a score after the fact. Between the
proposal and the score, however, the backtester has to decide when and how the
trade fills. Many studies use complete fills at the observed close without
discussing the choice.

That simplification would be harmless if execution costs moved every strategy
by roughly the same amount. There is little reason to assume that they do.
Orders cross spreads, wait, fill only in part, or move prices; strategies also
trade at different rates. Market-microstructure research has studied these
effects for decades \cite{kyle1985,almgrenchriss2001,bouchaud2009}, and practical
backtests routinely vary their cost assumptions. LLM-agent evaluations rarely
report whether their rankings survive the same exercise.

To measure this effect, twelve policies trade identical market paths under six
execution settings. Five obtain signals from LLMs and seven are classical
baselines. Market paths and decision frequencies are shared, so the comparison
changes execution without changing the information initially shown to an agent.

The largest rank change occurs in a short, high-volatility experiment with only
two assets. It is also the setting in which the leading policies are hardest to
distinguish. Wider universes have higher agreement point estimates, and
uncertainty is large. The finding is about the benchmark, not about which family
of policies trades better. Here ``LLM agent'' means a prompted signal policy:
the model produces one bounded signal per asset, while portfolio construction,
risk controls, and execution are shared. Agents with tools, roles, or persistent
memory are outside the main experiment (Section~\ref{sec:limitations}).

The paper's contribution is an empirical, paired comparison that holds market
paths and decision timing fixed while changing execution. We follow the
low-agreement case through turnover, universe size, horizon, anchor
initialization, and an empirical-price check. Run-level results and the code
used to build the tables and figures are publicly released
(Section~\ref{sec:reproducibility}).

\section{Related Work}

Financial language models \cite{wu2023,fingpt2023} and agent frameworks such as
TradingAgents \cite{tradingagents2024}, FinRobot \cite{finrobot2024}, and FinMem
\cite{yu2023} have made backtested trading a common evaluation task. Because
these systems differ in their use of tools, memory, and role-based reasoning, we
treat the model only as the source of a bounded signal and keep the downstream
portfolio and execution stack fixed.

Execution itself has a much older literature. Work on informed trading, optimal
liquidation, empirical impact, and transient order-book effects shows why costs
depend on both the order and the market \cite{kyle1985,almgrenchriss2001,
almgren2005,obizhaeva2013,gatheral2010,bouchaud2009}. Configurable transaction
costs are also standard in trading environments such as FinRL-Meta
\cite{liu2022finrlmeta}. Realized trading costs can differ substantially from
simple estimates and can change which strategies appear viable
\cite{frazzini2018,garleanu2013}.

The closest precedent is work on benchmark sensitivity. Backtest selection and
multiple testing can produce unstable conclusions
\cite{bailey2014,harvey2016,lopezdeprado2018}. LLM evaluations are also sensitive
to prompt wording and protocol choices \cite{sclar2024,mizrahi2024,liang2023},
and multi-step agent benchmarks introduce dependencies on the protocol and state
\cite{liu2023agentbench}. Finance-LLM surveys and broader benchmarks summarize
model performance \cite{liyaodong2023,nie2024,xie2024}. What has received less
attention is whether a trading leaderboard survives a change in its execution
assumptions. That is the question studied here.

\section{Execution Models}
\label{sec:ladder}

The experiments use the TradeArena testbed \cite{xue2026tradearena}. At each
decision, the harness converts a signal to target weights, applies the risk
gate, creates orders, simulates fills, and updates the portfolio.
Table~\ref{tab:execution} summarizes the six execution settings. E0 uses
immediate, complete fills. E1 adds latency, participation, impact, and the
possibility of partial fills or rejection. Each single-axis E2 setting starts
from E1 and changes one parameter; the harsh setting changes all three named
axes and market impact.

\begin{table}[t]
\caption{Execution settings. Latency is measured in bars; the participation cap
is the maximum share of bar volume; spread is the full quoted spread in basis
points; and $\kappa$ is the linear-impact coefficient. Every setting charges
1\,bp commission and 2\,bps base slippage. E1 and E2 also include the
intrabar-volatility term.}
\label{tab:execution}
\small
\setlength{\tabcolsep}{4pt}
\begin{tabular}{lrrrr}
\toprule
Setting & Latency & Cap & Spread & $\kappa$ \\
\midrule
E0 near-ideal      & 0 & --    & 0  & 0.00 \\
E1 default         & 1 & 5\%   & 0  & 0.15 \\
E2 spread          & 1 & 5\%   & 20 & 0.15 \\
E2 latency         & 3 & 5\%   & 0  & 0.15 \\
E2 participation   & 1 & 1\%   & 0  & 0.15 \\
E2 harsh           & 3 & 1\%   & 20 & 0.30 \\
\bottomrule
\end{tabular}
\end{table}

On the synthetic paths, an order receives
$q^{\mathrm{fill}}=\min(|q|,\rho V_t)$ shares, where $\rho$ is the
participation cap and $V_t$ is bar volume. Its execution price is
\[
p^{\mathrm{exec}}_t=p^{\mathrm{close}}_t\Big(1\pm\big[\tfrac{\text{spread}}{2}+s_0+\kappa\,\tfrac{q^{\mathrm{fill}}}{V_t}+\lambda\,\nu_t\big]\Big),
\]
The sign is positive for buys and negative for sells. We use $s_0=2$\,bps,
$\lambda=0.1$, and $\kappa=0.15$ at E1 or $0.30$ in the harsh setting. Here
$\nu_t=(\text{high}_t-\text{low}_t)/p^{\mathrm{close}}_t$. The quoted spread is
charged as a half-spread on each side, and the 1\,bp commission is applied
separately to notional. Any quantity above the participation cap is canceled
rather than queued. Latency $\ell$ moves an order to bar $t+\ell$.

Each run begins with \$100{,}000 in cash and prices near \$100. The seeded return
generator combines trend, seasonal, macro, news, and idiosyncratic terms. The
high-volatility regime scales the volatility, trend, and macro components by
$2.25$, $0.65$, and $1.4$; the jump/tail regime uses Student-$t_3$ shocks and
jumps. Volume is approximately $10^6(1+10|\varepsilon_t|)$ shares.

These settings are stress tests, not estimates of trading costs for a particular
market. The synthetic volume scale leaves the participation cap mostly slack for
a \$100{,}000 account. We repeat the deterministic experiment at \$30M, where the
cap binds (Section~\ref{sec:mechanism}), and sweep the parameters separately
(Section~\ref{sec:grid}). Every policy uses the same risk gate: a 0.35 per-name
cap, gross exposure of 1.0, and a five-bar drawdown kill switch at $-20\%$.

\section{Experimental Setup}
\label{sec:setup}

\textbf{Agents.} The seven deterministic baselines are buy-and-hold,
signal-weighted allocation, naive momentum, five-bar mean reversion, 12-bar risk
parity, 24-bar minimum variance, and a seeded random allocator. They are
long-only. All but buy-and-hold rebalance at every step. Per-name weights are
capped at 0.35, except minimum variance, for which we use 0.08 to prevent a
near-singular two-asset covariance estimate from concentrating the portfolio in
one asset. In the two-asset buy-and-hold portfolio, each asset receives 35\% and
the remainder stays in cash.

The five LLM policies use four provider-routed aliases (gpt-5.5,
gemini-3.1-pro, claude-opus-4.7, and glm-5) and one direct-API model
(deepseek-v4-pro). The routed aliases do not identify stable backend
checkpoints. We treat them as observations of the interfaces available during
collection, not as reproducible measurements of named model versions. The
direct API also uses an undated identifier, so it does not establish a fixed
backend version.

Each LLM receives recent OHLCV bars, portfolio state, and risk feedback in a
structured JSON prompt. It returns one signal in $[-1,1]$ per asset, not target
weights. A shared strategy converts those signals into weights before the common
risk and execution layers run. The prompt does not name the execution setting
or expose its parameters. The first decision is therefore identical across
settings. Later decisions may differ because earlier fills have changed the
portfolio state.

Responses are cached by
\texttt{provider:model:sha256(prompt)}, with timestamps removed from the prompt.
Provider caches are not redistributed. The public release
contains the run-level data used in the paper together with hashes for those
files.

\textbf{Market regimes.} A seeded simulator generates calm-trend,
high-volatility, and jump/tail paths. The generated paths cannot have appeared
verbatim in a model's training data, which reduces direct test-set memorization.
Section~\ref{sec:realmarket} adds a deterministic check on real OHLCV data; we
do not use that historical window to make claims about LLM performance.

\textbf{Repetition and pairing.} We use ten seeds in every
(regime, level, agent) cell. A seed fixes the market path, and every agent takes
12 decisions on that path at each execution level. The main matrix contains
$12\times3\times6\times10=2{,}160$ runs. Another 302 LLM runs support the
sampling-variance check, for 2,462 records and 14,424 provider calls in total.
The robustness experiments are counted separately. We compute Sharpe for each
run and then average across seeds within a cell.

\textbf{Statistics.} Within a cell, policies are ranked by mean Sharpe across
seeds. Run-level Sharpe is the mean per-step equity return divided by its
standard deviation and scaled by $\sqrt{252}$. The 12-step window makes these
values large, so they should be compared only within the experiment. We measure
agreement between execution levels with Kendall $\tau_b$ and top-3 Jaccard
overlap.

For friction fragility, we compare each policy $a$ with buy-and-hold $b$ using
the per-seed difference-in-differences
$(R^{E0}_{a,s}-R^{L}_{a,s})-(R^{E0}_{b,s}-R^{L}_{b,s})$. A positive value means
that $a$ loses more than the anchor when moving to stressed level $L$. We pool
the 30 paired regime-seed differences and use a two-sided sign-flip test: exact
enumeration for at most 16 values and 2,000 Monte Carlo assignments with a
plus-one correction otherwise. Benjamini--Hochberg adjustment covers the single
declared family of 11 non-anchor agents at E1 and the harsh setting. The other
execution levels and regime-specific summaries are exploratory. With only three
regime blocks, the latter cannot support a regime-level significance claim.
Repeated provider samples are averaged within seed before the comparison.

\section{Results}
\label{sec:results}

\subsection{Execution Assumptions Change the Ranking}

\begin{figure}[t]
\centering
\includegraphics[width=\linewidth]{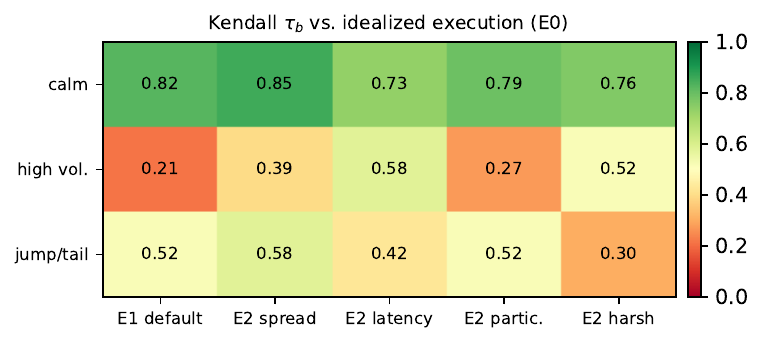}
\Description{A three-by-five heatmap of Kendall tau-b between idealized and
stressed execution rankings. The smallest value, 0.21, occurs for default
stress in the high-volatility regime; calm-regime values are all above 0.7.}
\caption{Kendall $\tau_b$ between the E0 ranking and each stressed ranking.
Rows are market regimes and columns are execution settings. Each cell contains
12 policies ranked by mean Sharpe over ten seeds; red cells indicate less
agreement and green cells more.}
\label{fig:heatmap}
\end{figure}

Figure~\ref{fig:heatmap} compares the E0 ranking with each stressed ranking.
The calm board is fairly stable, with $\tau_b$ between 0.73 and 0.85. The
jump/tail board is less stable (0.30--0.58). The largest change occurs in the
high-volatility E0--E1 comparison: $\tau_b=0.21$, with a seed-bootstrap 95\%
CI of $[0.00,0.52]$. The corresponding calm estimate is 0.82
($[0.48,0.94]$). These intervals are wide and overlap. In every stressed
high-volatility setting, the E0 and stressed top threes share only one policy
(Jaccard 0.2).

Nominally harsher settings do not always produce larger rank changes. Within
the high-volatility regime, E1 gives $\tau_b=0.21$ and the 1\% participation
setting gives 0.27. At this account size those settings are nearly equivalent;
the small difference comes from provider resampling rather than a binding cap
(Section~\ref{sec:mechanism}). The combined harsh setting gives 0.52, while the
single-axis latency and spread settings give 0.58 and 0.39. A monotone pattern
appears only when we vary a common parameter grid from liquid to illiquid
(Section~\ref{sec:grid}).

\subsection{Why Sharpe Moves More Than Total Return}

For a run, Sharpe is $S=\sqrt{252}\,\bar r/s_r$, calculated from 12 portfolio
returns. Execution changes the sequence $r_1,\ldots,r_{12}$: latency moves
exposure from one bar to another, partial fills reduce it, and trading costs are
deducted when a fill occurs. Total return largely summarizes the compounded
endpoint. Sharpe uses both the mean and dispersion of the path, so the same fill
can affect its numerator, its denominator, or both. This matters most on the
high-volatility paths, where moving exposure by one bar changes which return the
portfolio holds.

The difference is visible in the rankings. For total return, E0--E1 $\tau_b$ is
$0.97/0.70/0.61$ in calm, high-volatility, and jump/tail markets. For Sharpe it
is $0.82/0.21/0.52$. The $\sqrt{252}$ annualization factor is not responsible:
it is common to every policy and leaves their order unchanged. The uncertain
part is $s_r$, estimated here from only 12 returns. A fill that changes one or
two of them may have little effect on terminal wealth yet noticeably change the
sample volatility. That difference matters when several policies already have
similar scores.

This distinction changes how the result should be read. Total return records
where the portfolio finishes; Sharpe also records the path taken to get there.
We use Sharpe as the risk-adjusted ranking measure, but the return ranking makes
clear that the largest movement comes from estimating risk on a short window,
not from endpoint performance alone.

\subsection{Which Policies Move}

Table~\ref{tab:flip} shows where the high-volatility change comes from. Under
E0, the top five are all classical policies. Under E1, every LLM policy moves up
between two and seven places, and four enter the top five. The two weakest
classical policies remain at the bottom. Turnover differs, but not cleanly:
classical policies average 18--20 E1 fills and the LLM policies average 15--18,
with similar fill rates of 0.80--0.86. The confidence intervals overlap
substantially. What changes clearly is the ordering; the table does not separate
the trading ability of the two groups.

Some LLM policies even gain a small amount of Sharpe and total return from E0
to E1; deepseek-v4-pro moves from $+0.050$ to $+0.076$ total return. This need
not reflect better decisions. Frequent rebalancing is already costly in this
short experiment, and E1 suppresses some of those trades. A lower-turnover
policy can move up in relative terms even when execution becomes less favorable
for the board as a whole.

\begin{table}[t]
\caption{High-volatility ranking under E0 and E1, ordered by E0 rank. Values
are mean Sharpe over ten seeds with 95\% bootstrap CIs. $\Delta$ is the change
in rank from E0 to E1 ($+$ means a higher rank under E1). Sharpe magnitudes are
comparable within the table, not with annualized real-market values.}
\label{tab:flip}
\footnotesize
\setlength{\tabcolsep}{3.5pt}
\begin{tabular}{llcccc}
\toprule
Agent & Type & E0 Sharpe & \#$_{E0}$ & E1 Sharpe & $\Delta$ \\
\midrule
risk-parity        & cls & 11.07 {\scriptsize[8.8,13.6]} & 1 & 8.69 {\scriptsize[6.6,10.9]} & $-5$ \\
buy-and-hold       & cls & 10.95 {\scriptsize[9.0,12.9]} & 2 & 9.46 {\scriptsize[7.5,11.6]} & $-1$ \\
naive-momentum     & cls & 10.90 {\scriptsize[8.3,13.8]} & 3 & 8.26 {\scriptsize[6.4,10.1]} & $-6$ \\
minimum-variance   & cls & 10.07 {\scriptsize[8.0,12.2]} & 4 & 8.56 {\scriptsize[6.7,10.3]} & $-4$ \\
random             & cls &  9.35 {\scriptsize[7.3,11.7]} & 5 & 7.75 {\scriptsize[5.1,10.0]} & $-5$ \\
\textit{glm-5}            & llm & 9.22 {\scriptsize[6.3,12.0]} & 6 & 9.59 {\scriptsize[7.8,11.5]} & $+5$ \\
\textit{gpt-5.5}          & llm & 9.02 {\scriptsize[6.2,11.9]} & 7 & 9.06 {\scriptsize[7.3,10.8]} & $+2$ \\
\textit{claude-opus-4.7}  & llm & 9.02 {\scriptsize[6.1,11.9]} & 8 & 9.22 {\scriptsize[7.6,10.8]} & $+4$ \\
\textit{deepseek-v4-pro}  & llm & 8.19 {\scriptsize[5.6,10.8]} & 9 & 9.51 {\scriptsize[7.8,11.5]} & $+7$ \\
\textit{gemini-3.1-pro}   & llm & 7.71 {\scriptsize[4.6,10.8]} & 10 & 8.65 {\scriptsize[6.9,10.6]} & $+3$ \\
signal-weighted    & cls &  6.13 {\scriptsize[2.6,9.0]}  & 11 & 4.85 {\scriptsize[1.4,7.2]}  & $\;\;0$ \\
mean-reversion     & cls &  3.72 {\scriptsize[0.2,6.7]}  & 12 & 3.30 {\scriptsize[0.2,5.9]}  & $\;\;0$ \\
\bottomrule
\end{tabular}
\end{table}

\subsection{Ranking Changes Within Turnover Groups}
\label{sec:turnover}

We group the policies into terciles using realized E1 turnover and recompute
E0--E1 agreement within each group. The mean within-tercile $\tau_b$ is 0.63.
In the high-volatility regime, all three terciles give 0.67 even though the full
board gives 0.21. Much of the full-board movement lies between turnover groups,
although policies with similar turnover still reorder. This is a descriptive
check: each tercile contains only four policies, and E1 turnover is itself an
outcome of the experiment.

We also add a cost-aware risk-parity baseline that trades only after a weight
drifts by more than 0.05. Its mean E1 turnover falls from 20--21 fills to
14--17. It ranks first under high-volatility E0 and sixth under E1. With this
policy included, E0--E1 $\tau_b$ is 0.69, 0.23, and 0.51 across the three
regimes. The high-volatility and jump/tail values remain close to 0.21 and
0.52; the calm value falls from 0.82 to 0.69. Adding this less active baseline
does not remove the low-agreement case, although the extension has only one
provider sample per cell.

\subsection{Leaderboard Compression}
\label{sec:compression}

The amount of reordering is associated with how closely the leaders are packed.
Across the 15 regime-by-stress cells, the stressed-level Sharpe gap between rank
1 and rank 5 correlates with E0 agreement at $r=0.75$. Every cell with a top-five
gap below 0.6 has $\tau_b\le0.42$; calm cells have gaps of 1.4--2.2 and
$\tau_b>0.72$. These 15 cells cluster within three regimes, so we use the
correlation as a description rather than a test. It fits what is visible in the
high-volatility table: the leading confidence intervals overlap.

\begin{figure*}[t]
\centering
\includegraphics[width=\textwidth]{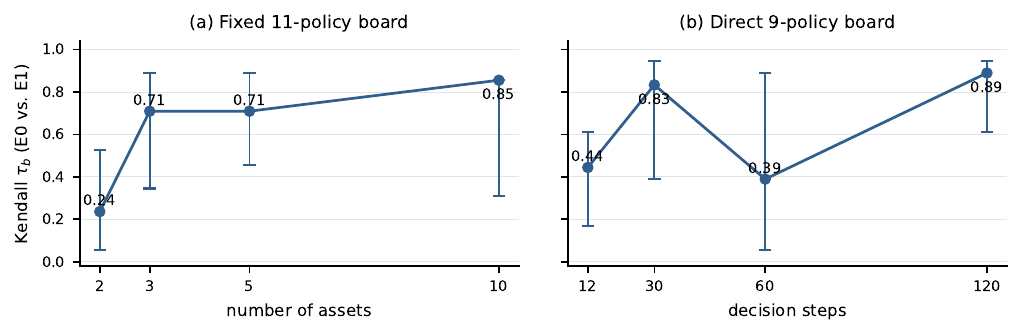}
\Description{Two line charts with seed-bootstrap confidence intervals. The
left chart shows Kendall tau-b rising as the universe grows from two to ten
assets on a fixed eleven-policy board. The right chart shows a non-monotone
relationship between Kendall tau-b and decision horizon on a nine-policy direct
board.}
\caption{High-volatility E0--E1 rank agreement under two robustness checks.
(a) The fixed 11-policy board varies universe size at 12 decisions. (b) The
direct nine-policy board varies decision horizon with two assets. Points are
Kendall $\tau_b$; bars are seed-bootstrap 95\% intervals from ten seeds.}
\label{fig:robustness}
\end{figure*}

Figure~\ref{fig:robustness}a keeps the same 11 policies at every universe size:
seven classical policies, deepseek-v4-pro, and three routed policies. E0--E1
$\tau_b$ rises from 0.24 with two assets to 0.71 with three and five, and 0.85
with ten. All four intervals are broad and overlap. We read the curve as a
boundary on the two-asset estimate, not as evidence of a monotone universe-size
effect.

The second panel changes the decision horizon instead of the universe. It uses
the reproducible nine-policy direct board, not the 12-policy headline board.
Agreement does not rise steadily with the horizon. It reaches 0.83 at 30 steps,
falls to 0.39 at 60, and rises to 0.89 at 120, from 0.44 at 12 steps. The
60-step interval, $[0.06,0.89]$, is especially wide; only the 120-step interval,
$[0.61,0.94]$, separates from the 12-step interval, $[0.17,0.61]$.

\subsection{The Buy-and-Hold Anchor}

The difference-in-differences comparison asks a separate question: whether a
policy loses more return than buy-and-hold when friction is added. At E1, no
policy is more fragile than the anchor; deepseek-v4-pro is 2.5 percentage points
less fragile ($q=0.029$). In the harsh setting all point estimates are negative.
Deepseek-v4-pro ($-4.0$ points) and mean reversion ($-4.5$ points) have
$q=0.005$. The signs are consistent across regimes, but three regime blocks do
not support a separate regime-level test.

The anchor starts from cash, so its measured loss includes the cost of building
the initial position. When the anchor starts with its target portfolio already
in place, the warm-minus-cold return difference is zero at E0 and 0.025 on
average in the harsh setting, reaching 0.046 in jump/tail. Part of the anchor's
apparent fragility is therefore an initialization effect. The full-board rank
comparison does not use this anchor.

We estimate sampling variance under independent errors with a common
within-seed variance. If $W$ is the mean sample variance of seeds with repeats,
$V$ is the sample variance of all $S$ seed means, and seed $s$ has $n_s$ samples,
the between-seed estimate is $B=\max\{0,V-W S^{-1}\sum_{s=1}^{S}n_s^{-1}\}$.
The estimated share is $W/(B+W)$; truncation and limited repeats make it
uncertain. Our summary uses the 14 cells with three samples at each of ten
seeds: the mean share is about 17\% (range below 0.001 to 0.53), with 0.14 in
the high-volatility E1 cell of the headline comparison. Four unevenly repeated
cells remain in the released table but are excluded from this summary; two
estimate $W$ from only one repeated seed. Averaging samples changes DiD estimates by at most 0.4 percentage
points. Only the deepseek-v4-pro E1 comparison crosses the threshold after
averaging, so we treat that result as marginal.

\begin{table}[t]
\caption{Friction-fragility difference-in-differences relative to buy-and-hold,
in percentage points of total return. Negative values mean less loss than the
anchor. $q$-values use seed-level sign flips and BH correction across 22 tests;
bold indicates $q<0.05$.}
\label{tab:did}
\small
\setlength{\tabcolsep}{4pt}
\begin{tabular}{llrrrr}
\toprule
 & & \multicolumn{2}{c}{E1 default} & \multicolumn{2}{c}{harsh corner} \\
\cmidrule(lr){3-4}\cmidrule(lr){5-6}
Agent & Type & DiD & $q$ & DiD & $q$ \\
\midrule
deepseek-v4-pro  & llm & \textbf{-2.46} & \textbf{.029} & \textbf{-3.97} & \textbf{.005} \\
mean-reversion   & cls & -0.68 & .535 & \textbf{-4.48} & \textbf{.005} \\
gemini-3.1-pro   & llm & -0.57 & .655 & -2.30 & .266 \\
claude-opus-4.7  & llm & -0.65 & .535 & -1.93 & .266 \\
glm-5            & llm & -0.54 & .535 & -1.70 & .410 \\
gpt-5.5          & llm & -0.37 & .677 & -1.42 & .448 \\
naive-momentum   & cls & -0.37 & .677 & -1.39 & .266 \\
random           & cls & +0.25 & .686 & -1.76 & .266 \\
signal-weighted  & cls & +0.14 & .789 & -1.31 & .295 \\
risk-parity      & cls & +0.58 & .448 & -0.96 & .266 \\
minimum-variance & cls & +0.52 & .535 & -0.95 & .266 \\
\bottomrule
\end{tabular}
\end{table}

\subsection{Which Frictions Bind}
\label{sec:mechanism}

At the \$100{,}000 account size, a position is at most about 350 shares per
asset, while synthetic bar volume is around one million shares. The 5\% and
1\% participation caps therefore allow roughly 50,000 and 10,000 shares. No
deterministic fill reaches either cap, realized participation stays below
0.1\%, and linear impact is well below one basis point. Eleven of the 12
policies produce identical records under the two cap settings. The remaining
difference disappears when the direct model is replayed from a fixed response
cache, confirming that it came from resampling.

E1 fill rates lie between 0.78 and 0.86 for both LLM and classical policies.
They fall to 0.37--0.54 in the harsh setting, mainly because of three-bar
latency. Similar fill rates do not imply similar rank changes. In high
volatility, latency lowers the fill rate to 0.48 but leaves $\tau_b=0.58$;
E1 leaves the fill rate near 0.83 but gives $\tau_b=0.21$. The spread setting
has the largest slippage and gives $\tau_b=0.39$. What matters is which policies
a cost penalizes, especially how often and how much they rebalance.

Consistent with this, a regression of E0--E1 return loss or rank shift on a
policy's own fill rate and slippage cost gives $R^2<0.02$ ($p>0.5$ for every
coefficient, $n=36$ cells). These diagnostics do not provide a reliable
after-the-fact correction for the leaderboard.

Finally, we repeat the deterministic experiment at \$30M. The participation
cap now binds on 33--95\% of fills, but high-volatility E0--E1 agreement remains
$\tau_b=0.81$ at both account sizes. For this policy set, per-trade costs and
turnover matter more than whether the cap binds.

\subsection{Parameter Sensitivity}
\label{sec:grid}

In the high-volatility regime, we also vary impact, participation, and latency
together along a liquid-to-illiquid grid. A liquid setting (impact 0.05, 10\%
participation, no latency) leaves the deterministic ranking unchanged
($\tau_b=1.0$). A typical-equity setting gives 0.81, and the illiquid end
(impact 0.30, 1\% participation, three-bar latency) gives 0.43. Unlike the
heterogeneous E2 settings, this controlled grid produces a monotone pattern.

\subsection{Real-Price Robustness}
\label{sec:realmarket}

We run the seven deterministic policies on real Yahoo OHLCV using the same
execution settings. The data cover eight liquid ETFs and bitcoin during a 2022
rates drawdown and a 2025 cross-asset window, with weekly adjusted closes, 12
decisions, and three offsets. Across both windows, E0-to-stress $\tau_b$ ranges
from 0.71 to 0.90. The top three change only under the spread setting (Jaccard
0.5; otherwise 1.0). This is milder than the synthetic two-asset result, but it
shows that deterministic rankings can also move on empirical price and volume
paths. It is not a historical test of the LLM policies.

\section{Discussion}
\label{sec:discussion}

The practical point is narrower than the familiar statement that execution
matters. In the high-volatility synthetic cell, one simulator choice changes
the apparent leaders from classical policies to LLM policies. The intervals
overlap, so neither ordering demonstrates that one group is better. Instead,
the comparison shows that a benchmark can turn an implementation choice into
its headline result.

A useful evaluation should state its fill rule, costs, latency, participation
limits, decision frequency, and anchor initialization. It should then rerun the
whole board under at least one plausible alternative and report rank agreement,
not only changes in return. Shared market paths and paired statistics make that
comparison more informative, and run-level records let others inspect it.

Subtracting an estimated cost after the run is not equivalent. A policy's own
fill rate and slippage explain almost none of its rank shift ($R^2<0.02$). The
shift depends on how the same friction acts across the full set of policies. A
leaderboard should therefore be rerun, rather than adjusted one score at a time.

\section{Limitations}
\label{sec:limitations}

Our strongest result is also the narrowest. The headline board has two assets,
12 decisions, and several policies with similar Sharpe estimates. In the fixed
11-policy universe experiment, high-volatility E0--E1 $\tau_b$ rises from 0.24
at two assets to 0.85 at ten. A same-seed deterministic ten-asset experiment
gives 0.79--0.93 across the five stressed settings. The value 0.21 is a result
for the compressed board, not a general effect size. We exclude a
provider-backed ten-asset E2 run because its E0 and E2 seeds do not match; a
fully paired LLM ladder at that size remains to be run.

The short horizon makes both Sharpe and terminal handling noisy. Pending orders
are canceled after 12 decisions, so latency $\ell$ can censor up to $\ell/12$
of late orders. We did not run the matched flush-versus-cancel comparison needed
to isolate that choice. The seed-bootstrap intervals for high-volatility and
calm E0--E1 agreement are wide and slightly overlap.

Longer horizons give a mixed pattern (Figure~\ref{fig:robustness}b). The
deterministic board remains between 0.71 and 0.91 from 12 to 120 steps, while
the direct board is non-monotone and imprecise. The still-available routed
subset gives $0.94,0.56,0.83$ at 30, 60, and 120 steps. Reordering can recur,
but the extreme 0.21 estimate belongs to a short, compressed board.

The LLM experiment is closed-loop. Once fills differ, a model sees a different
portfolio state and may change its next signal. The deterministic policies
provide a mechanical comparison because their targets are identical across
execution levels and the drawdown switch never fires ($0/840$ steps). For the
direct LLM board, replaying the E0 decisions under E1 gives $\tau_b=0.56$,
compared with 0.44 when the policies decide again. Mechanical execution changes
the ranking on its own. The additional difference is consistent with feedback
and fresh provider sampling, but these two $\tau_b$ values do not separate
their contributions.

Four of the five headline LLM policies are routed. None of the hosted API
identifiers binds the observations to an immutable backend version.
The frozen records preserve the analyzed snapshot, but those responses cannot
be collected again byte for byte. The conclusions apply to the provider
interfaces observed during collection and to prompted signal policies, not to
stable model versions or richer agents with tools, reflection, and multiple
roles.

Finally, synthetic paths support pairing and avoid training-data contamination,
but they do not establish live performance. The real-OHLCV experiment covers
only deterministic policies on nine ETF and crypto instruments. Running the
LLMs on the same historical window would introduce a possible memorization
confound. A post-training-cutoff forward test and quote-level fill replay remain
outside this study.

\section{Reproducibility}
\label{sec:reproducibility}

The code, execution settings, seeds, run-level outcomes, and analysis scripts
used in the paper are available at
{\hypersetup{pdfborder={0 0 1},urlbordercolor={0 0 1}}%
\url{https://github.com/weich97/execution-assumptions-rankings}}.
The release includes each
regime-by-level leaderboard, bootstrap intervals, and per-policy turnover, fill
rate, and slippage cost. The tables and figures can be rebuilt from the frozen
records without an API call, and the deterministic experiments can be rerun from
the start. Fresh LLM collection requires provider access; raw provider responses
and per-step trajectories are not included.

\section{Conclusion}

We asked whether a trading-agent leaderboard survives when the agents, market
paths, and performance metric stay fixed but execution changes. On the
compressed two-asset board, it often does not; the wider-universe checks have
higher agreement. The result concerns what the benchmark measures, not which
policy family trades better. Execution rules belong to the benchmark
specification and should be varied before small rank differences are interpreted
as differences in agent quality.

\section*{Acknowledgments}

Claude Code and OpenAI Codex were used during code implementation and review,
analysis-script checking, and manuscript revision. The author designed the
study, ran the experiments, inspected the records, checked the citations, and
takes responsibility for the paper.

\bibliographystyle{ACM-Reference-Format}
\bibliography{refs}

\end{document}